\pdfoutput=1
\documentclass{article}

\PassOptionsToPackage{numbers, compress}{natbib}
\usepackage[preprint]{neurips_2023}

\usepackage[utf8]{inputenc} 
\usepackage[T1]{fontenc}    
\usepackage{hyperref}       
\usepackage{url}            
\usepackage{booktabs}       
\usepackage{amsfonts}       
\usepackage{nicefrac}       
\usepackage{microtype}      
\usepackage{xcolor}         
\usepackage{xspace}
\usepackage{pifont}
\usepackage{graphicx}
\usepackage{caption}
\usepackage{tcolorbox}
\usepackage{listings}
\definecolor{lean-keyword}{HTML}{80a1c1}
\definecolor{lean-parentheses}{HTML}{b48dad}
\definecolor{lean-number}{HTML}{ffd602}
\definecolor{lean-suggestions}{HTML}{1a9467}

\lstdefinelanguage{Lean4}{
  morekeywords={by, example, Type},
  keywordstyle=[1]\color{lean-keyword},
  sensitive=true,
  literate=%
    {0}{{{\color{lean-number}0}}}1
    {1}{{{\color{lean-number}1}}}1
    {2}{{{\color{lean-number}2}}}1
    {3}{{{\color{lean-number}3}}}1
    {4}{{{\color{lean-number}4}}}1
    {5}{{{\color{lean-number}5}}}1
    {6}{{{\color{lean-number}6}}}1
    {7}{{{\color{lean-number}7}}}1
    {8}{{{\color{lean-number}8}}}1
    {9}{{{\color{lean-number}9}}}1
    {(}{{{\color{lean-parentheses}(}}}1
    {)}{{{\color{lean-parentheses})}}}1
    {\{}{{{\color{lean-parentheses}\{}}}1
    {\}}{{{\color{lean-parentheses})\}}}}1
}

\newcommand{\llmstep}{\textsc{llmstep}\xspace}

\title{\llmstep: LLM proofstep suggestions in Lean}

\author{%
  Sean Welleck \\
  University of Washington\\
  Carnegie Mellon University \\
  \And 
  Rahul Saha \\ 
  \texttt{rsaha@alumni.princeton.edu} \\
}

\begin{document}

\maketitle

\begin{abstract}
  We present \llmstep, a tool for integrating a language model into the Lean proof assistant.
  \llmstep is a Lean 4 tactic that sends a user's proof state to a server hosting a language model. 
  The language model generates suggestions, which are checked in Lean and displayed to a user in their development environment. We provide a baseline language model, along with code for fine-tuning and evaluation to support further development.
  We provide server implementations that run on CPU, a CUDA GPU, or a Google Colab notebook, as a step towards fast, effective language model suggestions for any user.
\end{abstract}

\section{Introduction}
Interactive proof assistants such as Lean  \cite{de2015lean}, Isabelle  \cite{wenzel2008isabelle}, and Coq  \cite{coq} enable the verification of mathematics and software using specialized programming languages \cite{Avigad,Ringer2019}.
The emerging area of neural theorem proving integrates neural language models with interactive proof assistants \cite{first2023baldur,polu2020generative,polu2022formal,yang2023leandojo,ntptutorial}.
Doing so can be mutually beneficial: proof assistants provide correctness guarantees on language model outputs, while language models may help make proof assistants easier to use.
A fundamental part of proof development is determining which step to take next at each state of a proof (i.e., which  \textit{tactic} to use).
Therefore, a tool that suggests useful next steps within a user's development environment could significantly ease proof development.

We present \llmstep, a tool for suggesting proof steps (i.e., tactics) with a language model in the Lean proof assistant (\autoref{fig:llmstep}).
\llmstep is a Lean 4 tactic that sends a user's proof state to a server hosting a language model. 
The language model generates suggestions, which are checked in Lean and displayed to a user in their development environment.
\llmstep is agnostic to the choice of language model, learning framework, and evaluation framework. 
We provide a baseline language model and example code for fine-tuning and evaluation.
The baseline language model is fine-tuned for a standard tactic-prediction task \cite{han2022proof}, and outperforms recent open-source tactic-prediction models \cite{yang2023leandojo}.
Finally, \llmstep supports several runtimes, with servers that run on CPU, a GPU, or in a Google Colab notebook, as a step towards fast, powerful language model suggestions for any user.\footnote{\url{https://github.com/wellecks/llmstep}}

\section{Related Work}

Automatically generating proof steps with language models  is an active area of research (e.g., \cite{polu2020generative,ntptutorial}). 
Closest to our work is the \texttt{gpt-f} tactic from  \cite{han2022proof}, which generates suggestions in Lean 3 by calling a (now disabled) Open-AI API.
\llmstep is inspired by the idea of language-model based suggestions, but differs in several ways: (1) \llmstep is built with open-source components, and can run on a user's own device. (2) \llmstep supports prefixed and checked suggestions, detailed below. (3)  \llmstep is in Lean 4, requiring an implementation using Lean 4 metaprogramming.  (4) \llmstep provides code for fine-tuning and evaluating future models.
Recently, after the release of \llmstep, LeanInfer \cite{leaninfer} provides the ability to run supported language models on CPU through Lean's Foreign Function Interface (FFI). 
Similar to \llmstep it provides tactic suggestions, and uses \llmstep's utilities to check, format, and display its suggestions. 
\llmstep additionally supports prefixed suggestions, offers fast GPU inference, and is agnostic to the language model implementation.
Proofster \cite{agrawal23icse} is a related tool offering machine-learning based proof synthesis in Coq via a web interface, while \llmstep offers Lean 4 language-model tactic suggestions in the development environment.

\section{Approach}

\llmstep is called by writing \texttt{llmstep <prefix>} within a proof, which returns suggestions that start with \texttt{<prefix>} (for instance, $\texttt{llmstep ""}$ or $\texttt{llmstep "exact"}$).
\llmstep uses Lean to check whether each suggestion is valid and/or completes the proof.
The suggestions are displayed in the Lean 4 VS Code Infoview, which is a standard interface used in proof development.
A user can click a suggestion, which places it in the proof.
The proof is either complete, or it transitions to the next state and the user continues writing the proof.
We detail \llmstep's implementation below.

\subsection{Implementation}

\begin{figure}
    \centering
    \includegraphics[width=\textwidth]{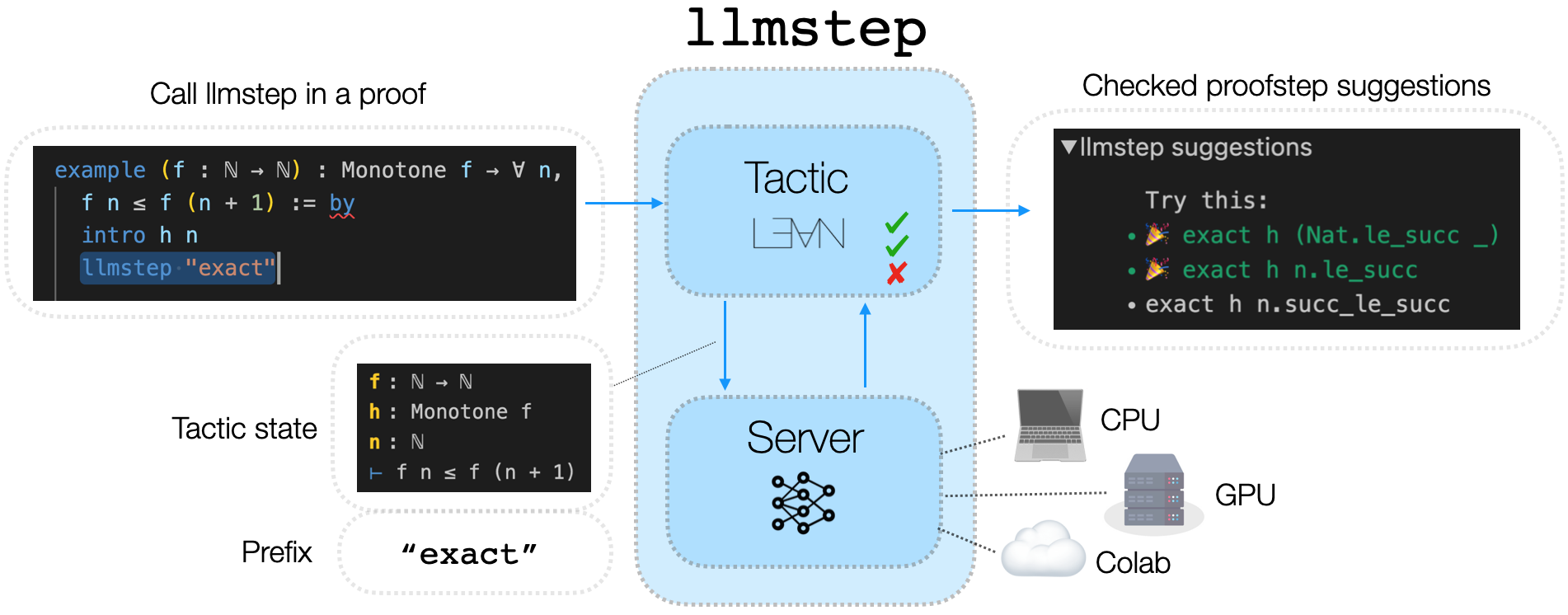}
    \caption{\llmstep is a tool for receiving proofstep suggestions from a language model. A user calls \llmstep in a proof (\textit{left}), which sends the current state of the proof and a prefix to a server. 
    A language model generates proofstep suggestions which are checked in Lean and shown to the user (\textit{right}). \llmstep supports a variety of compute environments, with servers that run on CPU, CUDA GPU, or Google Colab. 
    \llmstep is implemented as a Lean 4 metaprogram with Python servers.
    }
    \label{fig:llmstep}
\end{figure}

\llmstep consists of three parts: (1) a Lean \textit{tactic}, (2) a \textit{language model}, (3) a \textit{server}.
\paragraph{Lean tactic.}
Writing a proof can be seen as a sequential process $(x_1,y_1),(x_2,y_2),\ldots$ of \textit{states} $x_t$ and \textit{tactics} $y_t$. 
A state contains what is left to prove (the \textit{goal}), and available information (the \textit{hypotheses}).
A \textit{tactic} transitions the proof to a new state.
If the state contains no remaining goals, the proof is complete. 
Concretely, a user applies tactics by writing Lean code, Lean keeps track of the state, and the development environment shows the state and the written code.

\llmstep is itself a tactic.
\llmstep takes a prefix as an argument, i.e., a sequence of tokens that will start the suggested tactics. For instance, \texttt{"intr"} would lead to suggested tactics that start with \texttt{"intr"}, such as \texttt{intro h x}.
\llmstep sends the current state $x_t$ and the prefix to a server, \llmstep receives suggestions in response, and
\llmstep checks each suggestion using Lean.
Namely, a checked suggestion is \textit{valid} if applying the tactic leads to a state with no errors and at least one goal.
A tactic is \textit{complete} if it leads to a state with no errors and no goals.
Otherwise the tactic is \textit{invalid}.
\llmstep displays complete, valid, and invalid tactic suggestions using different colors.

\paragraph{Language model.} \llmstep uses a language model to predict the next tactic given the current tactic state. 
While the language model can be arbitrary, one approach is to use a model that has been fine-tuned on (state, next-tactic) examples. 
For instance, the default language model in \llmstep is fine-tuned on sequences of the form:    
\begin{tcolorbox}[sharp corners=all]
\begin{center}
\texttt{
[GOAL]tactic-state[PROOFSTEP]next-tactic<|endoftext|>
}    
\end{center}
\end{tcolorbox}
This format corresponds to the proofstep objective described in  \cite{han2022proof}.

By default, \llmstep uses a {\small \texttt{Pythia 2.8b}} language model \cite{biderman2023PythiaAS} fine-tuned on (state, next-tactic) examples extracted from Lean Mathlib  \cite{mathlib} via the LeanDojo Benchmark 4 dataset \cite{yang2023leandojo,yang_kaiyu_2023_8040110}. The {\small \texttt{Pythia 2.8b}} model is publicly available on Huggingface 
(\href{https://huggingface.co/wellecks/llmstep-mathlib4-pythia2.8b}{link}).
\llmstep is agnostic to the language model implementation, and includes direct support for ReProver  \cite{yang2023leandojo} and  other Huggingface models.

\textbf{Server.} \llmstep uses a server to handle requests from the Lean tactic and host the language model. 
The server queries the language model and relays responses back to the Lean tactic. 
The server is the key computational bottleneck in \llmstep, as it hosts a (possibly large) language model. 
\llmstep supports a variety of compute constraints, with server implementations that run on CPU, a CUDA GPU, or a  Google Colab notebook with GPU, as well as a server with fast inference via vLLM \cite{kwon2023efficient}.

\textbf{Usage.}
First, the user starts a server based on their hardware. Currently, servers can run on CPU, CUDA GPU, or Google Colab notebooks.
Second, the user imports \llmstep as a Lean 4 package. 
Third,  the user calls \llmstep by writing \texttt{llmstep} \texttt{<prefix>}, which returns suggestions that start with the prefix passed to \llmstep. The suggestions are displayed in the Infoview.

\section{Evaluation}

First, we benchmark the default language model's utility for providing suggestions via proof search--i.e., attempting to fully prove  theorems using the language model and a search algorithm.

\textbf{Proof search.}
Proof search requires a search algorithm and a method for interacting with Lean.
We use best-first search, and provide a self-contained implementation
with 
LeanDojo  \cite{yang2023leandojo} interaction. 

Best-first search is parameterized by the maximum number of generated tactics, defined as the number of attempts  $\times$ expansion size per iteration $\times$  maximum iterations, subject to a timeout.
We use a 10 minute timeout as in  \cite{yang2023leandojo}, and use beam search with expansion size 32 based on memory constraints.
We compare the Pythia model to ReProver  \cite{yang2023leandojo} without retrieval.
To equal the 64 expansions used in  \cite{yang2023leandojo}, we also report results with a second attempt of 32 samples (i.e., 2$\times$32).
We report mathlib4-test from  \cite{yang2023leandojo} and miniF2F-test from  \cite{thakur2023ALA}, since miniF2F without retrieval was not available in  \cite{yang2023leandojo}.

\begin{table}[h]
\centering
\begin{tabular}{lccccc}
\toprule
\textbf{Model}  & \textbf{Search} & \textbf{mathlib4-valid} & \textbf{mathlib4-test} & \textbf{miniF2F-valid} & \textbf{miniF2F-test}\\
\midrule
ReProver &$1\times 64$ & -- & 48.6\% & -- & 22.1\% (54/244)\\
 Pythia 2.8b&$1\times 32$ & 48.8\% & 47.6\% & 26.2\% & 27.9\% (68/244)\\
 Pythia 2.8b&$2\times 32$ & 51.2\% & 50.1\% & 26.2\% & 27.9\% (68/244)\\
\bottomrule
\end{tabular}
\vspace{2mm}\caption{Proofsearch on the Lean Dojo random evaluation splits and miniF2F.
}
\label{table:benchmark}
\vspace{-3mm}
\end{table}

We validate the Pythia model in \autoref{table:benchmark}, finding that it can exceed the number of closed theorems by ReProver on Lean Dojo Benchmark 4 and miniF2F.
Note that \llmstep supports suggestions from either model. 
The ReProver model is particularly useful on CPU, as we discuss next.

\begin{figure}[htpb]
    \centering
    \includegraphics[width=200px]{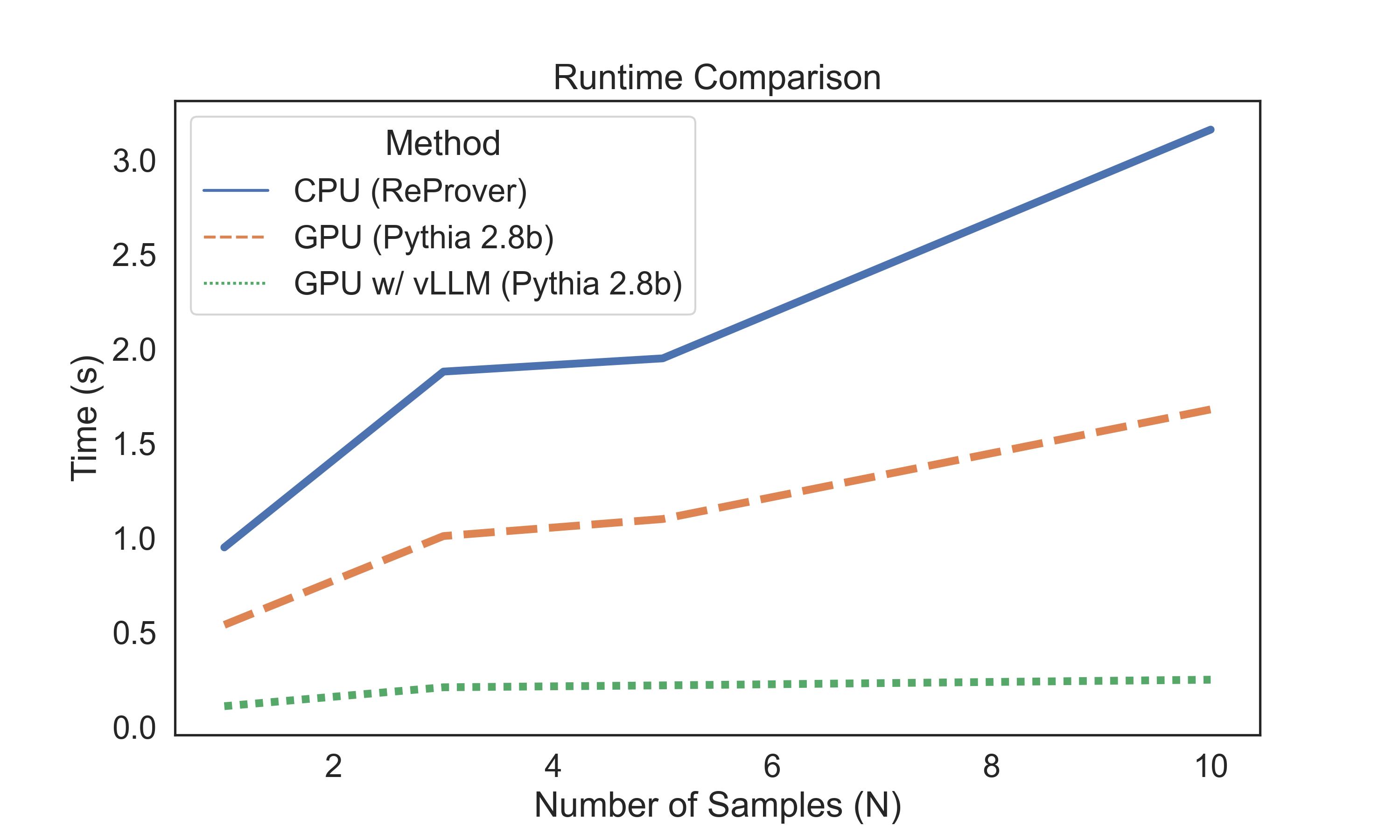}
    \caption{Runtime comparison of different compute environments and models supported by \llmstep. 
    }
    \label{fig:runtime}
\end{figure}

\textbf{Runtime.}
We tested the runtime of \llmstep with different compute environments and hardware. The experiments were done on a set of 17 examples, each containing a tactic state and a prefix. 
For each example, we measured the time for the server to return $N$ suggestions, using \textsc{llmstep} with the {\small \texttt{Pythia 2.8b}} and ReProver ({\small \texttt{leandojo-lean4-tacgen-byt5-small}}) language models.

\begin{table}[h]
\centering
\vspace{2mm}
\begin{tabular}{lllll}
\toprule
\textbf{Compute} & \textbf{Model}& \textbf{Hardware} & \textbf{N}&\textbf{Time} \\
\midrule
 CPU & Pythia 2.8b & MB Pro 2.6 GHz 6-Core Intel Core i7  & 1&  8.39s\\
 CPU & ReProver & MB Pro 2.6 GHz 6-Core Intel Core i7  & 1&  0.95s\\
 \midrule
 CPU & Pythia 2.8b & MB Pro 2.6 GHz 6-Core Intel Core i7  & 10&  49.10s \\
 CPU & ReProver & MB Pro 2.6 GHz 6-Core Intel Core i7  & 10&  3.16s\\
 \midrule
 Colab GPU & Pythia 2.8b & NVIDIA T4 GPU & 1& 1.01s\\  
 Colab GPU & Pythia 2.8b & NVIDIA T4 GPU & 10& 1.68s\\  
 \midrule
 GPU w/ vLLM & Pythia 2.8b & NVIDIA GTX 1080Ti  & 1&0.11s \\ 
 GPU w/ vLLM & Pythia 2.8b & NVIDIA GTX 1080Ti  & 10& 0.25s\\ 
\bottomrule
\end{tabular}
\vspace{2mm}\caption{Runtime experiments done with different compute resources and hardware, using the {\small \texttt{Pythia 2.8b}} and Reprover models in \textsc{llmstep}. 
\textit{N} refers to the number of generated suggestions.
The experiments report average runtime across 17 examples.}
\label{tab:runtime}
\vspace{-3mm}
\end{table}

As shown in \autoref{tab:runtime}, \llmstep with GPU-based inference can yield suggestions in 1 second or less, with vLLM inference approaching real-time (0.11s).
As shown in \autoref{fig:runtime}, vLLM remains fast as the number of suggestions increases.
Note that vLLM does not support the model architecture used in ReProver.
However, on CPU the ReProver model is much faster due to its small parameter count.
Therefore, we suggest using Pythia when a GPU is available, and Reprover on CPU.

\textbf{Qualitative examples.} Figures \ref{fig:llmstep} and \ref{fig:example-from-dataset-1} show example suggestions made by $\llmstep$.

\begin{figure}[hpb]
  \small
  \begin{tcolorbox}[minipage,arc=0pt, standard jigsaw, opacityback=0, outer arc=0pt, size=normal, sharp corners=all]
    \begin{tabular}{l l}
    \textbf{Input (User)} &
\begin{lstlisting}[language=Lean4]
example : $R \subseteq S \rightarrow S \subseteq T \rightarrow R \subseteq T$ := by
    llmstep ""
\end{lstlisting} \\
\midrule
\textbf{Suggestion (\llmstep)} &
\begin{lstlisting}
Lean Infoview
  Try This:
    * <@ \color{lean-suggestions}{exact subset\_trans} @>
    * <@ \color{lean-suggestions}{exact Set.Subset.trans} @>
    * <@ \color{lean-suggestions}{tauto} @>
    * <@ \color{lean-keyword}{intro}@>
    * <@ \color{lean-keyword}{intros h1 h2}@>
\end{lstlisting} \\
    \end{tabular}
  \end{tcolorbox} 
  \caption{In this example, the user inputs an empty prefix. \llmstep returns five suggestions. The first three suggestions each finish the proof, and the last two are valid.}
  \label{fig:example-from-dataset-1}
\end{figure}

\vspace{-3mm}
\section{Conclusion}

In this paper, we present \llmstep, a tool designed to make it easy for a Lean 4 user to obtain tactic suggestions from a language model. 
In addition, we provide a fine-tuned language model that achieves strong performance on mathlib and miniF2F. 
Active areas of work include fast CPU inference, improved models  \cite{azerbayev2023llemma}, and tasks beyond tactic prediction.
We hope that \llmstep's simple, model-agnostic recipe opens up new research avenues on generative tools for formalized mathematics. 

\section{Acknowledgements}
We thank Mario Carneiro, Zhangir Azerbayev, and Scott Morrison  for valuable guidance and feedback.

\bibliography{neurips_2023}
\bibliographystyle{plainnat}


\end{document}